\documentclass{article}
\usepackage{ijcai24}

\usepackage{times}
\usepackage{soul}
\usepackage{url}
\usepackage[hidelinks]{hyperref}
\usepackage[utf8]{inputenc}
\usepackage[small]{caption}
\usepackage{amsthm}
\usepackage[switch]{lineno}

\usepackage{booktabs}       
\usepackage{amsfonts}       
\usepackage{nicefrac}       
\usepackage{microtype}      
\usepackage{xcolor}         
\usepackage{xspace}
\usepackage{dsfont}
\usepackage{amsmath,amssymb}
\usepackage[ruled,norelsize]{algorithm2e}
\usepackage{multirow}
\usepackage{booktabs}
\usepackage{graphicx}
\usepackage{subcaption}
\usepackage{natbib}
\usepackage{float}
\hypersetup{
   colorlinks=true,
   linkcolor=blue,
   filecolor=blue,      
   urlcolor=blue,
   citecolor=black,
}

\newtheorem{definition}{Definition}[section]

\newif\ifcomments
\commentstrue

\newcommand{\ttt}[1]{\texttt{#1}}

\usepackage{amsmath,amsfonts,bm}

\newcommand{\eq}[1]{Eq.(\ref{#1})}
\newcommand{\anB}[1]{\langle #1 \rangle}
\newcommand{\brB}[1]{\left( #1 \right)}

\def\rve{{\mathbf{e}}}

\def\rmX{{\mathbf{X}}}

\def\vt{{\bm{t}}}

\def\vx{{\bm{x}}}

\def\mE{{\bm{E}}}

\def\mT{{\bm{T}}}

\def\mX{{\bm{X}}}

\DeclareMathAlphabet{\mathsfit}{\encodingdefault}{\sfdefault}{m}{sl}
\SetMathAlphabet{\mathsfit}{bold}{\encodingdefault}{\sfdefault}{bx}{n}

\def\cD{{\mathcal{D}}}
\def\cE{{\mathcal{E}}}

\def\cG{{\mathcal{G}}}

\def\cL{{\mathcal{L}}}

\def\cP{{\mathcal{P}}}
\def\cQ{{\mathcal{Q}}}

\def\cT{{\mathcal{T}}}

\def\cX{{\mathcal{X}}}

\SetKwInput{KwInit}{Init}
\SetKwInput{KwInput}{Input}
\newcommand{\randombvisit}{%
\begin{algorithm}[t]
\caption{Multi-start Random B-walk}
\SetAlgoLined

\KwInput{Graph $\cG$, start nodes $\rmX^0=\{x_1, x_2,...\}$}

\KwInit{Path $R=[]$, active nodes $\rmX=\{\}$}

Set all start nodes as active $\rmX \gets \rmX^0$

\While{$|\rmX| > 0$}{
    Get all viable edges $\mE \gets B\text{-}Connected(\rmX;\cG)$ 

    Sample $\rve := P(\vx_h, x_t, \vt) \sim Uniform(\mE)$

    Set $x_t$ as active $\rmX \gets \rmX \cup \{x_t\}$
    
    \For{$x_h \in \vx_h$}{
        \If{$AllOutEdgesVisited(x_h)$}{
            Set $x_h$ inactive $\rmX \gets \rmX/\{x_h\}$
        }
    }

    Add the edge to path $R \gets R + [e]$
}

Return $R$

\label{algo:randombvisit}
\end{algorithm}}

\newcommand{\randomtimewalk}{%
\begin{algorithm}[t]
\caption{Path-consistency for temporal relation generalization.}
\SetAlgoLined

\KwInput{Path $R=[\rve^1,...,\rve^T]$}
\KwInit{Temporal relation table $B[\cdot,\cdot] = \emptyset$}

\For{$(\rve_1, \rve_2) \in R \times R$}{

    Update the temporal relation $B[\rve_1, \rve_2] \gets TempRel(\rve_1, \rve_2)$

    \For{$\rve_3 \in R/\{\rve_1, \rve_2\}$}{

        Resolve path consistency $B[\rve_1, \rve_3], B[\rve_2, \rve_3] \gets PC3(\rve_1, \rve_2, \rve_3)$
        
    }

}

Return B

\label{algo:randombtime}
\end{algorithm}}

\title{Temporal Inductive Logic Reasoning over Hypergraphs}

\author{
    Yuan Yang\textsuperscript{1} \and
    Siheng Xiong\textsuperscript{1} \and
    Ali Payani\textsuperscript{2} \and    
    James C. Kerce\textsuperscript{1} \And 
    Faramarz Fekri\textsuperscript{1}
    \affiliations
    Georgia Institute of Technology\textsuperscript{1},     Cisco\textsuperscript{2}
    \emails
    \{yyang754@, 
        sxiong45@,
        clayton.kerce@gtri.,
        faramarz.fekri@ece.\}gatech.edu, 
    apayani@cisco.com
}

\pubnote{ \qquad \qquad \quad \    Proceedings of the Thirty-Third International Joint Conference on Artificial Intelligence (IJCAI-24)}

\begin{document}

\maketitle

\begin{abstract}
Inductive logic reasoning is a fundamental task in graph analysis, which aims to generalize patterns from data. This task has been extensively studied for traditional graph representations, such as knowledge graphs (KGs), using techniques like \textit{inductive logic programming} (ILP). 
Existing ILP methods assume learning from KGs with static facts and binary relations.
Beyond KGs, graph structures are widely present in other applications such as procedural instructions, scene graphs, and program executions.
While ILP is beneficial for these applications, applying it to those graphs is nontrivial: they are more complex than KGs, which usually involve timestamps and n-ary relations, effectively a type of hypergraph with temporal events.
In this work, we propose \textit{temporal inductive logic reasoning} (TILR), an ILP method that reasons on temporal hypergraphs. 
To enable hypergraph reasoning, we introduce the \textit{multi-start random B-walk}, a novel graph traversal method for hypergraphs.
By combining it with a path-consistency algorithm, TILR learns logic rules by generalizing from both temporal and relational data.
To address the lack of hypergraph benchmarks, we create and release two temporal hypergraph datasets: YouCook2-HG and nuScenes-HG.
Experiments on these benchmarks demonstrate that TILR achieves superior reasoning capability over various strong baselines. 
\end{abstract}

\section{Introduction}

The task of inductive reasoning concerns generalizing concepts or patterns from data. This task is studied extensively in knowledge graphs (KG)s where techniques such as \textit{inductive logic programming} (ILP) are proposed. A typical knowledge graph such as FB15K~\citep{toutanova2015observed} and WN18~\citep{bordes2013translating} represents commonsense knowledge as a set of nodes and edges, where entities are the nodes and the facts are represented as the edges that connect the entities. For example, \texttt{Father(Bob, Amy)} is a fact stating that ``\texttt{Bob} is the \texttt{Father} of \texttt{Amy}''; this is represented as an edge of type \texttt{Father} connecting the two entities \texttt{Bob} and \texttt{Amy}. Many ILP techniques are proposed to reason on the KGs that learn first-order logic rules from the graph. Learning these rules is beneficial as they are interpretable and data efficient~\citep{yang2020learn}. It has applications such as biomedical research, semantic search, data integration and fraud detection.

\begin{figure*}[t]
    \centering
    \includegraphics[trim={0 7.5cm 9.5cm 0},clip,width=0.8\linewidth]{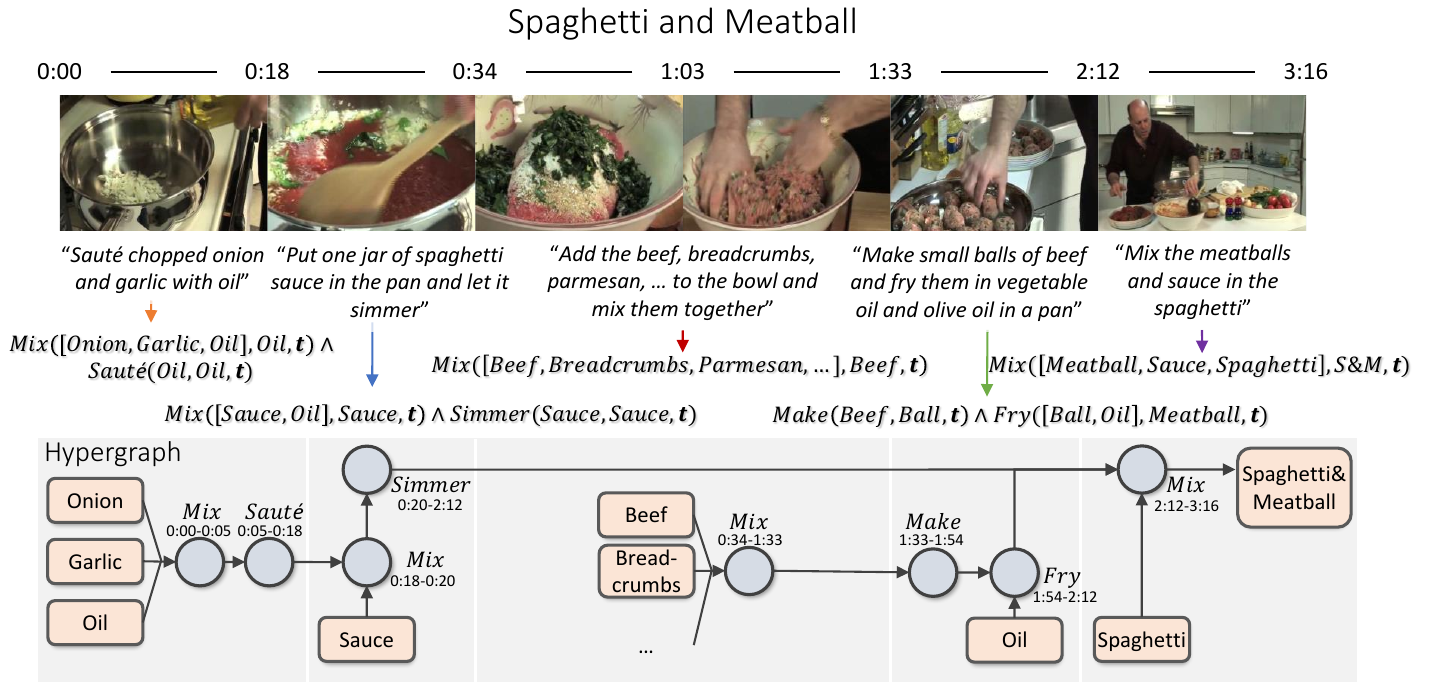}
    \caption{Temporal hypergraph representation of video instructions for making \textit{spaghetti $\&$ meatballs}.}
    \label{fig:example-graph}
\end{figure*}

Apart from KGs, the graph is widely used in other applications to represent structured data, such as temporal events that happened among a set of entities~\citep{Boschee2015icews,leetaru2013gdelt}; the cooking instruction of a video (Figure~\ref{fig:example-graph}); the abstract syntax tree (AST) of a program; and a scene graph from autonomous driving sensors~\citep{caesar2020nuscenes}. Learning explicit rules from these graphs is also beneficial. However, these graphs are more complex than standard graphs, and therefore, existing ILP methods are not readily applicable. Specifically, 
(1) for temporal data such as video, the facts or \textit{events} are labeled with time intervals indicating their start and end times. While some temporal KGs such as ICEWS~\citep{Boschee2015icews} also have time labels, events are labeled with only a single time point. This is less expressive than time intervals as it cannot characterize events with durations. For example, ``cook the soup while cutting the lettuce''. 
(2) Many events require n-ary relations, for example ``mixing onion, garlic, and oil together''. 
Such a relation corresponds to an edge that connects to more than two nodes, and a graph with such edges is a \textit{hypergraph}. 
While it is possible to convert hypergraphs to graphs with various techniques such as clique expansion, the conversions are lossy and can lead to an exponential number of edges. 

In this work, we extend traditional ILP methods to hypergraphs whose events are labeled with time intervals. 
To this end, we first formally define this representation, namely \textit{temporal hypergraph}. 
Then, we discuss the random walk (RW) algorithm on a hypergraph, as RW is the fundamental mechanism of a family of widely used ILP methods, that is the backward-chaining methods.
We revisit the notion of \textit{B-connectivity} on hypergraphs and propose the \textit{multi-start random B-walk} algorithm that explores the hypergraph given a set of starting points.
To generalize the temporal relation, we incorporate Allen's interval algebra~\citep{allen1983maintaining} and characterize events with the interval operators such as \texttt{Before} and \texttt{After}, and learn temporal relations by resolving the time constraints with \textit{path consistency algorithm}.
Our contributions are as follows:
\begin{itemize}
    \item We introduce the \textit{temporal hypergraph}, a graph that supports n-ary relations and temporal events with time intervals. We show that this is a natural and expressive representation for many applications.
    \item We propose TILR, an ILP method that learns to generalize on both temporal and higher-order relational data in hypergraphs. This is realized with a novel random B-walk algorithm and a time-constraint propagation algorithm.
    \item We release two novel temporal hypergraph datasets \textbf{YouCook2-HG} and \textbf{nuScenes-HG} \href{https://github.com/gblackout/TILR}{here}, which is created from YouCook2 cooking recipe dataset and nuScenes autonomous driving dataset~\citep{caesar2020nuscenes}. We show that TILR outperforms existing embedding-based and ILP baselines significantly on these two benchmarks.
\end{itemize}
To the best of our knowledge,
\textbf{YouCook2-HG} and \textbf{nuScenes-HG} are the first benchmarks dedicated to temporal hypergraphs, and TILR is the first framework that addresses the ILP problem on hypergraphs with time interval labels.

\section{Related Work}
\textbf{ILP methods}. Many ILP methods are proposed for inductive logic reasoning on graphs. These methods are categorized into two types. Forward-chaining methods~\citep{galarraga2015fast,evans2018learning,payani2019inductive} learn by searching in the rule space. It supports inferring difficult tasks but suffers from exponential complexity and does not scale to large graphs. On the other hand, backward-chaining methods~\citep{campero2018logical,yang2020learn,yang2017differentiable} learn rules by searching in the graph space, while the rule is usually limited to a chain-like path in the graph, this leads to better scalability. 
For temporal hypergraphs, none of the existing ILP methods are readily applicable.
In this work, we propose the ILP method for temporal hypergraphs under the backward-chaining paradigm.

\textbf{Reasoning on temporal knowledge graphs}. Temporal reasoning has been studied extensively on the temporal knowledge graphs (KG)s~\citep{nguyen2018continuous,ronca2018stream,trivedi2017know,pareja2020evolvegcn}.
Different from the temporal hypergraphs proposed in this work, these temporal KGs typically consist of only unary and binary relations with a single time point tag.
Recent methods such as TLogic~\citep{Liu2022tlogic} were proposed for solving ILP on these graphs, but \textit{time point algebra} is limited as it cannot represent temporal relations of events with duration. 
A more expressive representation is proposed in \textit{time interval algebra} with Allen's interval algebra~\citep{allen1983maintaining} being the representative schema. 
TILP~\citep{xiong2022tilp} adopted such an algebra and improved ILP methods for ordinary graphs.
In this work, we show how to incorporate this algebra for learning temporal relations on hypergraphs.


\section{Temporal Hypergraphs}

An ordinary graph consists of a set of \textit{triples} of the form $P(x_h, x_t)$, where $P$ is the predicate and $x_h$ and $x_t$ are the \textit{head} and \textit{tail} entities.
However, many real-world data cannot be fully captured by this formalism.
We use Figure~\ref{fig:example-graph} as a running example to show this.
Consider an instruction video that teaches how to make \ttt{Spaghetti$\&$Meatball} using ingredients such as \ttt{Onion}, \ttt{Oil}, and \ttt{Spaghetti}, which are processed to make the dish in a step-by-step manner.
Such information naturally implies a graphical structure, but two aspects cannot be captured by an ordinary graph
1) \textbf{Higher-order relations}.
Actions such as \ttt{MixInto} involve multiple entities, which requires n-ary relations to represent, which calls for a \textit{hypergraph} representation where the edges are now \textit{hyperedges}.
Note that, however, many techniques convert hypergraphs into ordinary graphs\citep{carletti2020random}, but it is proven that such conversion will lose the higher-order information\citep{chitra2019random,hayashi2020hypergraph}. 
Therefore, we focus on developing native algorithms for hypergraphs.
2) \textbf{Time intervals}. To properly represent events such as ``Saute the ingredients from 0:00 to 0:18'', a time interval $\anB{t_s, t_e}$ marking the start and the end time is needed.

\textbf{Temporal Hypergraph}.
To address the above challenges, we propose the \textit{temporal hypergraph}.
Formally, a temporal hypergraph consists of the following components:
\textit{Entity}: $x\in\cX$ is the unique entity in the temporal data. \textit{Predicate}: $P\in\cP$ is the temporal predicate. \textit{Event}: an event is defined as $\rve := P\brB{\vx_h, \vx_t, \vt}$, where $\vx_h, \vx_t$ are the \textit{head} entities and \textit{tail} entities\footnote{Note that in some work such as~\citep{gallo1993directed}, \textit{head} and \textit{tail} are used in a reversed manner}, and $\vt:=\anB{t_s, t_e}$ consists of the start and end time of an event. 
Note that an event $\rve$ is a natural extension to the triple fact in an ordinary graph: a triple $P(x_h, x_t)$ represents an edge with label $P$ that starts from a single entity $x_h$ to another entity $x_t$; 
similarly, $\rve := P\brB{\vx_h, \vx_t, \vt}$ represents a \textit{hyperedge} with label $P$ that starts from a set of entities $\vx_h$ and ends at a set of entities $\vx_t$ during a period of time $\vt:=\anB{t_s, t_e}$.
For example, $\ttt{MixInto}([\ttt{Ball}, \ttt{Sauce}, \ttt{Spaghetti}],$ $ \ttt{S\&M}, \anB{\text{2:12,3:16}})$ in Figure~\ref{fig:example-graph} represents an event of ``mixing three ingredients, i.e., \ttt{Meatball}, \ttt{Sauce} and \ttt{Spaghetti}, into a a dish of \ttt{Spaghetti$\&$Meatball} from 2:12 to 3:16''.

\begin{definition}[Temporal Hypergraph]
\label{def:temporal-hypergraph}
A temporal hypergraph $\cG$ is defined as $\cG := \anB{\cX, \cP, \cE}$, where $\cX$ and $\cP$ are the space of entities and space of predicates respectively, and $\cE = \{\rve_1, ..., \rve_n\}$ is a collection of events, where each event represents a hyperedge $\rve := P\brB{\vx_h, \vx_t, \vt}$ with timestamps $\vt:=\anB{t_s, t_e}$.
\end{definition}

A temporal hypergraph can represent many temporal data that are usually beyond the capacity of ordinary graphs, for example, representing the instructions/procedure to repair/build something, recipes in the video data shown in Figure~\ref{fig:example-graph}, or detecting the behavior patterns in autonomous driving logs.

\section{Temporal Inductive Logic Reasoning}

Temporal Inductive Logic Reasoning (TILR) concerns solving the ILP problem on the temporal hypergraph, which involves learning first-order logic (FOL) rules that generalize over the patterns in the graph. 

\textbf{First-Order Logic}.
A FOL rule consists of (i) a set of predicates defined in $\cP$, (ii) a set of logical variables such as $X$ and $Y$, and (iii) logical operators $\{\land, \lor, \neg \}$. For example, a FOL rule
$
    R: \texttt{GrandFatherOf}(X, X') \gets 
    \texttt{FatherOf}(X, Y) \land 
    \texttt{MotherOf}(Y, X') 
$
has predicates \texttt{GrandFatherOf}, \texttt{MotherOf} and \texttt{FatherOf}. Terms such as $\texttt{FatherOf}(X, Y)$ are referred to as \textbf{atom}s, which correspond to the predicates that apply to the logical variables. 
Each atom can be seen as a Boolean function.
For example, for the binary relation \ttt{FatherOf}, the atom is a mapping $\cX \times \cX \mapsto \{0, 1\}$. This function can be evaluated by \textbf{instantiating} the logical variables such as $X$ into the object in $\cX$. 
For example, let 
$\cX=\{\text{Amy},\text{Bob}\}$,
we can evaluate $\texttt{FatherOf}(\text{Bob}/X, \text{Amy}/Y)$ by instantiating $X$ and $Y$ into Bob and Amy respectively. 
This yields \texttt{True} if ``Bob is the father of Amy''. 
The outputs of atoms are combined using logical operations $\{\land, \lor, \neg \}$ and the \textit{imply} operation $a \gets b$ is equivalent to $a \lor \neg b$. 
Thus, when all variables are instantiated, the rule will produce an output as the specified combinations of those from the atoms. 

\textbf{Inductive Logic Programming (ILP)}.
Given a set of positive and negative queries $\cQ_+$ and $\cQ_-$, the ILP problem concerns learning a set of logic rules that predict (or entail) positive ones and do not predict the negative ones.
For example, summarizing a recipe for making \textit{Spaghetti $\&$ Meatball} from a set of videos (Figure~\ref{fig:example-graph}) is an ILP task. 
Let $\cG_1,...,\cG_n$ be the temporal graphs of the videos, ILP learns a logic rule that predicts the positive video labels $\cQ_+ = \{S\&M(\cG_1), ..., S\&M(\cG_n)\}$ and does not predict the negative video labels $\cQ_- = \{S\&M(\cG_i)|i\neq1,...,n\}$. 
Similarly, ILP can be used to answer event queries, 
where we learn rules to predict positive events $\cQ_+=\{\rve_1,...,\rve_n\}$, and not the negative events, which is essentially the link predicate task performed on the temporal hypergraph.

We solve the ILP problem via the backward-chaining approach: given a query $q$, one seeks to answer the query by finding a relation path in the graph that entails the
query~\citep{lao2011random,yang2017differentiable,yang2020learn}.
In an ordinary graph with a triple query $q=P(x^0, x^n)$, this relation path from $x^0$ to $x^n$ is represented as a chain-like rule
\begin{align}\label{eq:chain-like-rule-family}
    P(X^0, X^n) \gets P^1(X^0, X^1) \land ... \land P^n(X^{n-1}, X^n),
\end{align}
where $X^0, X^n$ are variables to be instantiated into the query entities $x^0$ and $x^n$, and $X^1,...,X^{n-1}$ are variables to be instantiated to the entities that exist along the relation path $x^0 \xrightarrow{P^{1}} x^1 \xrightarrow{P^2} ... \xrightarrow{P^n} x^n$.

\textbf{Temporal Inductive Logic Reasoning}. 
We extend this chain-like rule family for temporal hypergraphs to generalize over both the higher-order relational data and the temporal data.
Formally, we learn logic rules of the following form:
\begin{align}
\label{eq:temporal-chain-like-rule-family}
    P(\mX^0, \mX^n, \mT) & \gets P^1(\mX^0, \mX^1, \mT^1) \;\psi^1\; P^2(\mX^1, \mX^2, \mT^2) \nonumber \\ 
    & \;\psi^2\; ... \;\psi^{n-1}\; P^n(\mX^{n-1}, \mX^n, \mT^{n}).
\end{align}
Similar to~\eq{eq:chain-like-rule-family}, $\mX^0, \mX^n$ are variables of the head and tail entity sets, and $\mX^1,...,\mX^{n-1}$ are variables for those entities along the relation path which can now be in higher-order. On the other hand, we introduce the temporal operator $\psi \in \{\texttt{BEFORE}, \texttt{EQUAL}, \texttt{MEETS}, ...\}$ from Allen's interval algebra~\citep{allen1983maintaining} (details in \S\ref{sec:temporal-relation-generalization}), which is a widely accepted formalism for characterizing the temporal relations between time intervals. With this operator, a logic rule can now generalize over temporal data: similar to logical variables, $\mT, \mT^1, ..., \mT^{n}$ are the variables for the timestamps and can be instantiated into values $\vt, \vt^1, ...,  \vt^{n-1}$, and their temporal relations can be computed by the operators which yield \ttt{True} or \ttt{False} in a similar way as logical operators.



\begin{definition}[Temporal Inductive Logic Reasoning]
\label{def:tilr-objective}
Given a temporal hypergraph $\cG$, and a set of positive and negative queries $\cQ_+$ and $\cQ_-$, find a model $f(q;\cG)$ that maps the query $q$ into a logic rule $R$ such that
\begin{align*}
    R(q) = \mathds{1}[q\in\cQ_+],\; \text{for}\; q\in\cQ,
\end{align*}
where $\cQ = \cQ_+ \cup \cQ_-$ and $\mathds{1}[\cdot]$ is an indicator function.
\end{definition}

For an event query $q=\rve:=P(\vx^0, \vx^n, \vt)$, $R(\vx^0, \vx^n, \vt)$ denotes a logic rule with head predicate $P$ that entails $\rve$ is positive given the entities and the time interval. Similarly, for a graph label query $q=P(\cG)$, $R(\cG)$ denotes a logic rule with head predicate $P$ that entails if $\cG$ belongs to class $P$ given the entire graph.
However, learning the rule family~\eq{eq:temporal-chain-like-rule-family} on the temporal hypergraph is nontrivial and one needs to address two challenges: 
\textbf{(C1)}. How to traverse and learn higher-order relations on hypergraph via random walk? 
\textbf{(C2)} How to generalize over the temporal data while walking on the graph? We address these challenges in the following section.

\subsection{Random Walk on Temporal Hypergraphs}

In an ordinary graph, chain-like rules~\eq{eq:chain-like-rule-family} are typically learned via personalized graph random walk~\citep{lao2011random,lao2010relational}. 
Given a start node $x_h$, a single run of the random walk involves repeating the following two steps: 
(i) sample an out-edge of the current node $e \sim OutEdges(x_h)$, uniformly at random; 
(ii) move to the tail node $x_t$ and update $x_h \gets x_t$. 
Given a maximum length $n$, this yields a sample relation path, and by running random walks multiple times, one collects a set of relation paths $R_1, ..., R_n$, which can then be used in various differentiable models to learn the desired chain-like rules (details in \S\ref{sec:learning-model}).




\textbf{B-graph and B-connectivity}.
For temporal hypergraphs, however, while there exist many random walk techniques for hypergraphs~\citep{chitra2019random,chan2018spectral,li2020quadratic}, none of the existing work is designed for ILP and concerns an important property, i.e., the B-connectivity.
To see this, consider an event with \ttt{Mix} relation in the recipe example in Figure~\ref{fig:example-graph}, i.e., $\ttt{Mix}([\ttt{Onion}, \ttt{Garlic},$ $\ttt{Oil}], \ttt{Oil}, \vt)$. This is a hyperedge with 3 head nodes and 1 tail node. A fundamental issue arises in developing traversal algorithms for exploring the graph through this edge, that is, ``\textit{can we reach the tail node if we haven't visited all the head nodes?}''. Intuitively, the answer is no, as we consider the actual transformational nature of the event: ``\textit{we \ttt{Mix} the ingredients only when we have collected all three of them}''.

\begin{figure}[t]
    \centering
    \includegraphics[trim={0 4.8cm 15.5cm 0},clip,width=0.75\linewidth]{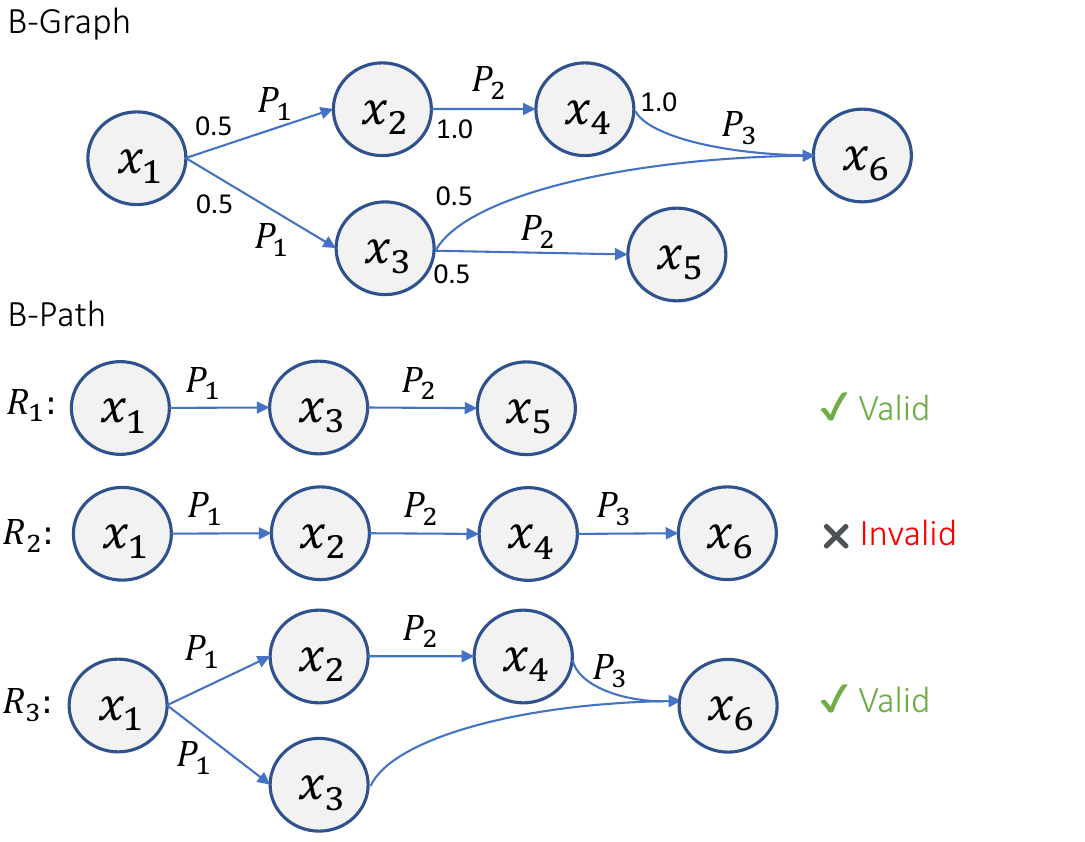}
    \caption{Example B-graph and B-paths.}
    \label{fig:b-graph-examples}
    \vspace{-.3cm}
\end{figure}

Specifically, let $\texttt{Mix}(\vx_h, x_t, \vt)$ be a \ttt{Mix} event with $\vx_h$ denotes the ingredients and $x_t$ (since $\text{dim}(\vx_t)=1$) denotes the mixed output.
Suppose at the $i$th step of a random walk, $\vx^{i}$ is the set of nodes that we have visited.
Then, there exists a natural constraint that ``$x_t$ is reachable via \texttt{Mix} if and only if $\vx_h \subseteq \vx^{i}$'', that is, the tail nodes cannot be reached until all the head nodes are reached. This property is referred to as \textit{B-connectivity}~\citep{gallo1993directed}. 
This problem can be further formulated with the notion of \textit{B-graph} which is a specific subset of directed hypergraphs.
Formally, let $P(\vx_h, \vx_t, \vt)$ be a hyperedge of predicate $P$. A hyperedge is a \textit{B-edge} if $\text{dim}(\vx_h)\geq1$ and $\text{dim}(\vx_t)=1$, meaning $P$ is a many-to-one relation. 
Nodes are \textit{B-connected}, if there exists a path consisting of B-edges that connects them, and the path is referred to as a \textit{B-path}. 
A hypergraph with B-edges is a \textit{B-graph}. Figure~\ref{fig:b-graph-examples} shows an example hypergraph with 3 paths, where $R_1$ and $R_3$ are valid B-paths and $R_2$ is the invalid one since it does not visit $x_3$ before $x_6$.
More discussion at \S\ref{app:hypergraph-discussion}.

\randombvisit

\textbf{Multi-start Random B-walk}. 
We propose a novel random walk method on the B-graph, i.e., the multi-start random B-walk (MRBW) to address challenge \textbf{(C1)}. The MRBW has two unique properties compared to the traditional random walk: 
1) it maintains a set of \textit{active nodes} $\rmX$, which keeps track of the nodes that have been visited but still have unvisited out-edges; 
2) for every hop, it samples an edge uniformly from all edges that are B-connected to $\rmX$, and then updates the active nodes accordingly.
Algorithm~\ref{algo:randombvisit} shows how to sample a B-path via MRBW. The function $B\text{-}Connected(\mX;\cG)$ collects all nodes that are B-connected to the active nodes $\mX$ and the corresponding B-edges $\mE$. Then, it randomly chooses an edge $\rve$ and hops to its tail node $x_t$. After each hop, the function $AllOutEdgesVisited(x_h)$ checks if all out-edges of each $x_h\in\vx_h$ have been visited and will remove it from the active nodes if true.

Specifically, to sample logic rules~\eq{eq:temporal-chain-like-rule-family}, we perform MRBW over B-paths with predicates sequence $P^1...P^n$.
One important feature to characterize the likelihood of such a path is the distribution of random walk given the specified predicates~\citep{lao2011random}.
Formally, let $\vx^{i-1}$ be the nodes reached at $i-1$th step, and $P^{i}$ be the predicate by which transition is performed at $i$th step.
If it is an ordinary graph, then $\vx^{i-1}=x^{i-1}$
and we have path probability to $x^i$ as
\begin{equation*}
    p(x^i | x^{i-1}; P^i) = 
        \frac{N_{P^i}(x^{i-1}, x^i)}{|N_{P^i}(x^{i-1}, \cdot)|} \cdot 
        p(x^{i-1} | x^{i-2}; P^{i-1}),
\end{equation*}
where $N_{P^i}$ is an indicator function that checks if $x^i$ and $x^{i-1}$ are connected by edge of type $P^i$ and $\cdot$ means arbitrary nodes.
For a hypergraph where 
$\text{dim}(\vx^{i-1}), \text{dim}(\vx^{i}) \geq 1$,
we extend the above, such that for
$\forall x^i \in \vx^i$ we have
\begin{equation}\label{eq:rw-prob-many-to-one}
    p_{i-1}^{RW}(x^i) =  
        \sum_{\vx_h \in \vx^{i-1}}
        \frac{N_{P^i}(\vx_h, x^i)}{|N_{P^i}(\vx_h, \cdot)|}
        \prod_{x_h\in\vx_h}
        p^{RW}_{i-2}(x_h),
\end{equation}
where $p^{RW}_{i-1}(\cdot)$ denotes $p(\cdot|\vx^{i-1}, P^i)$.~\eq{eq:rw-prob-many-to-one} specifies that for every head nodes $\vx_h\in\vx^{i-1}$ that are connected to $x^i$, the probability of them to land on $x^i$ is the product of $p^{RW}_{i-2}(x_h)$ for every individual node, and $p_{i-1}^{RW}(x^i)$ is the sum of probabilities of all such head nodes divided by their out degrees.
Finally, for path distribution over $\vx_i$ we have
\begin{equation}\label{eq:MRBW-transition-prob}
    p_{i-1}^{RW}(\vx^i) = 
        \prod_{x^i\in\vx^i} 
        p_{i-1}^{RW}(x^i).
\end{equation}
With~\eq{eq:MRBW-transition-prob}, we can now compute the likelihood of MRBW over a specified path $P^1...P^n$ for any logic rules of the form~\eq{eq:temporal-chain-like-rule-family}.
In \S\ref{sec:learning-model}, we build the differentiable model that utilizes this as the feature.



\subsection{Temporal Relation Generalization}~\label{sec:temporal-relation-generalization}

Here, we address challenge \textbf{(C2)} and incorporate Allen's Interval Algebra~\citep{allen1983maintaining} for generalizing over temporal data.
Allen's Interval Algebra consists of 6 asymmetric interval relations \{\ttt{BEFORE}, \ttt{DURING}, \ttt{MEETS}, \ttt{OVERLAPS}, \ttt{STARTS}, \ttt{FINISHES}\}, the corresponding reverse relations \{\ttt{AFTER}, \ttt{CONTAINS}, \ttt{MET-BY}, \ttt{OVERLAPPED-BY}, \ttt{STARTED-BY}, \ttt{FINISHED-BY}\}, and one symmetric relation \ttt{EQUAL}. Similar to the logical operators, each interval relation is a unique mapping of timestamps to the Boolean value \ttt{True} or \ttt{False}. For example, for two intervals $\vt,  \vt'$, \ttt{BEFORE} checks if $\vt$ ends earlier than where $\vt$ starts, e.g., $\anB{0:00,0:03} \;\ttt{BEFORE}\; \anB{0:04,0:10} = \ttt{True}$. All 13 relations are self-explanatory and the complete list of the operations is provided in~\citep{allen1983maintaining}.

TILR incorporates interval algebra into the logic rule. Recall the temporal rule family~\eq{eq:temporal-chain-like-rule-family}, instead of treating temporal relations as a separate operation from conjunction $\land$. we consider them as composite conjunction of the following form
\begin{align*}
    & P^i(\mX^{i-1}, \mX^i, \mT^i) \;\psi^i\; P^{i+1}(\mX^i, \mX^{i+1}, \mT^{i+1}) = \\
    & P^i(\mX^{i-1}, \mX^i) \land P^{i+1}(\mX^i, \mX^{i+1}) \land 
    \psi^i\brB{\mT^i, \mT^{i+1}},
\end{align*}
where $\psi^i \in \{\texttt{BEFORE}, \texttt{EQUAL}, \texttt{MEETS}, ...\}$. 
To generalize the temporal relations from the data, one keeps track of the set of applicable temporal relations for each pair of events in the rule. Whenever a positive query is matched by the rule, one updates the temporal relations to satisfy the time constraints posed in the corresponding subgraph of the query. 
We incorporate and modify the path-consistency (PC) algorithm~\citep{allen1983maintaining} for MRBW.
Algorithm~\ref{algo:randombtime} shows the process of resolving the temporal relations for a given path $R$. 
For every pair of events $(\rve_1, \rve_2)$ in the path, one first obtains the pairwise temporal relation via function $TempRel(\rve_1, \rve_2)$ which returns the satisfied relation.
Then, it checks the path consistency between the three edges of the path. This is done with $PC3(\rve_1, \rve_2, \rve_3)$ which is the PC3 algorithm.


\randomtimewalk

\subsection{Differentiable TILR for hypergraph reasoning}
\label{sec:learning-model}

Through MRBW and the PC algorithm, one can reliably traverse the temporal hypergraph and sample reasoning path, enabling us to develop models for solving the ILP task.
Here, we propose $\text{TILR}_{\bm{\theta}}$, a differentiable ILP model that reasons on the temporal hypergraph.

Let $R$ be a temporal relation B-path,
$\vx^0 \xrightarrow{P^{1}} ... \xrightarrow{P^n} \vx^n$ that traverses from $\vx^0$ to $\vx^n$ via relation $P^{1}...P^{n}$ and under the temporal constraints $\psi^1,...,\psi^{n-1}$.
We seek to model $R$ with a distribution $p_R$ as the path and temporal constrained random walk. 
Let $\cT^i = \{\vt | P^i(\vx_h, \vx_t, \vt) \in \cG\}$ be the space of time intervals in graph $\cG$ with respect to predicate $P^i$ at $i-$th step.
Given two such spaces $\cT^{i-1}, \cT^i$ , the space of time interval pairs that satisfy temporal constraint $\psi^{i-1}$ is computed as
$
\psi^{i-1}\brB{\cT^{i-1}, \cT^i} = 
    \{
        \anB{\vt^{i-1}, \vt^i} | 
            \psi^{i-1}\brB{\vt^{i-1}, \vt^i} = \ttt{True}; 
            \vt^{i-1}\in\cT^{i-1}, \vt^i\in\cT^{i} 
    \},
$
which represents the set of interval pairs where $\psi^{i-1}$ yields \ttt{True}.
Finally, the probability of traversing from step $i-1$ to step $i$ via under $\psi^{i-1}$ can be computed as
\begin{equation}\label{eq:PC-prob}
p(P^i|P^{i-1}) = 
    |\psi^{i-1}\brB{\cT^{i-1}, \cT^i}| / 
    |\psi^{i-1}\brB{\cT^{i-1}, \cdot}|,
\end{equation}
where $\cdot$ indicates arbitrary spaces in $\cG$.
We can now define distribution $p_R$ over $R$ with~\eq{eq:MRBW-transition-prob} and~\eq{eq:PC-prob} in a recursive fashion: 
\begin{equation}
    p_R(\vx^i | \vx^0) = p_{i-1}^{RW}(\vx^i) \cdot p(P^i|P^{i-1}).
\end{equation}
Recall Definition~\ref{def:tilr-objective}, to answer a query with respect to $\vx^0$ to $\vx^n$, we sample a set of paths $R_1 ... R_L$ and learn a parameterized model
\begin{equation}\label{eq:score-function}
    f_{\bm{\theta}}(q; \cG) = \sum_{j=1}^L \theta^p_j \sum_{i=1}^n \theta^s_i \cdot p_{R_j}(\vx^i | \vx^0),
\end{equation}
where $\bm{\theta}^p$ are the weights for choosing the appropriate paths and $\bm{\theta}^s$ are the weights for combining the path features over $p_{R_j}(\vx^i | \vx^0)$ the $n$ steps respectively.
To train such a model, we collect queries $\cD = \{q_i\}$ from the training split of the hypergraph $\cG$ by sampling the events and corresponding paths.
We optimize the model by minimizing a cross-entropy loss
\begin{equation*}
    \cL(\bm{\theta}; \cG) = 
        \sum_{q\in\cD} 
        y_i \log p(q | \cG) + 
        (1-y_i) \log (1 - p(q | \cG)), 
\end{equation*}
where $y_i\in{0,1}$ indicates if the query is positive, 
and $p(q | \cG) = 1 / (1 + e^{-f_{\bm{\theta}}(q; \cG)})$ is the conditional probability of $q$ computed from~\eq{eq:score-function} with sigmoid function.


\section{Experiment}

We conduct experiments to answer the following questions:
\textbf{(Q1)} can temporal hypergraphs preserve more information over ordinary graphs such that they lead to better performance in real-world tasks?
\textbf{(Q2)} Can TILR generalize better and interpretable logic rules from hypergraphs than traditional ILP methods?
\textbf{(Q3)} How do MRBW, PC algorithm, and parametrized model contribute to the performance of TILR?

To do so, we create two temporal hypergraph datasets \textbf{YouCook2-HG} and \textbf{nuScenes-HG} and evaluate TILR on the reasoning tasks of recipe summarization and driver behavior explanation.

\subsection{Recipe summarization}

YouCook2 cooking recipe dataset~\citep{zhou2018towards} consists of instructional videos of 89 cooking recipes such as \textit{spaghetti $\&$ meatballs}. Each recipe has 22 videos and each video is annotated by a sequence of natural language sentences that describe the procedure steps, as shown in Figure~\ref{fig:example-graph}.
The dataset serves as a challenging benchmark for evaluating the generalizability of both temporal and relational data: 
instructions of the same class can have varied procedures and ingredients.
For example, one can make \textit{BLT sandwich} by first putting the lettuce on the bread and then the ham, or in the reserve order.
This requires the ILP method to learn generalized temporal relations for the temporal events.

\begin{table}[H]
\centering
\resizebox{\linewidth}{!}{%
\begin{tabular}{@{}lccccc@{}}
\toprule
\multicolumn{2}{l}{\multirow{2}{*}{Predicate Type}} & \multirow{2}{*}{\# Predicates} & \multicolumn{2}{c}{\# Facts} & \multirow{2}{*}{Examples} \\
\multicolumn{2}{l}{}                                &                                & Per Graph       & Total      &                           \\ \midrule
\multirow{3}{*}{Relation}          & Unary          & 208                            & 58              & 10426      & Cut, fry                  \\
                                   & Binary         & 51                             & 29              & 5213       & Put                       \\
                                   & N-ary          & 1                              & 69              & 12417      & MixInto                   \\ \midrule
\multicolumn{2}{c}{Class}                           & 1136                           & 257             & 45881      & Oil, Oinon, Noodle        \\ \bottomrule
\end{tabular}%
}
\caption{Stats. of the hypergraph benchmark \textbf{YouCook2-HG}.}
\label{tab:recipe-stats}
\end{table}

\textbf{YouCook2-HG construction}. 
We construct the temporal hypergraph from the instruction sentences. For each clip, we run NLTK tools and extract the nouns and verbs from the sentences. 
We collect verbs as temporal relations. 
After pre-processing and grouping synonyms, there are 208 unary relations, 51 binary relations, and 1 n-ary relation. There are many possible n-ary relations in the raw data that describe the same action, such as ``put together'' and ``stir together''.
Given the low frequency of each data, we consider merging them into a single n-ary relation, that is \texttt{MixInto}. 
Statistics are shown in Table~\ref{tab:recipe-stats}. 

\textbf{Clique-expansion graphs}.
To evaluate \textbf{(Q1)} and \textbf{(Q2)}, an ordinary graph version of the \textbf{YouCook2-HG} needs to be created such that we can evaluate TILR on both the hypergraphs and the ordinary graphs with traditional baselines.
To do so, we use the clique expansion algorithm to convert the hypergraph to an ordinary graph: for every hyperedge
$\rve := P\brB{\vx_h, \vx_t, \anB{t_s, t_e}} \in \cE$, it creates edges for each pair of head and tail nodes
$\cE_\text{ord} = \{P(x_h, x_t, t_s) | x_h \in \vx_h, x_t \in \vx_t \}$.
Note that, for time intervals, we drop the end time $t_e$ and use $t_s$ as the time point label. 
This way, the clique-expansion graphs share the same structure as those traditional temporal graphs~\citep{Boschee2015icews}.


\textbf{Task}.
The goal of recipe summarization is to generalize a structured procedure for each distinct recipe from the hypergraph data as that in Figure~\ref{fig:example-graph}. We formalize this problem as a \textit{graph classification} task.
Formally, Let $P(\cG)$ be the label of a hypergraph $\cG$, where $P$ denotes its recipe type label (e.g. \textit{BLT sandwich}).
Let $P^*$ be the target type for which we want to summarize, then the positive set of $P^*$ is the set of all graphs with the same label, and the negative set is the rest of the graphs with different labels
\begin{align}\label{eq:recipe-background-pos-negs}
    \cQ_+ = \{P(\cG) | P = P^*\}, &&
    \cQ_- = \{P(\cG) | P \neq P^*\}.    
\end{align}
Recall the objective in Definition~\ref{def:tilr-objective}, 
we learn rules for recipe type $P^*$ that can classify graphs in $\cQ_+$ as positive and those in $\cQ_-$ as negative.
Furthermore, we also constrain the temporal rules to cover the full timespan of the graphs, ensuring rules fully represent the underlying recipe (details in \S\ref{app:exp-details}).

\textbf{Methods and baselines}.
We evaluate three modes of TILR with increasing complexity, so that it serves as an ablation study on the proposed modules: 
1) TILR: this mode performs vanilla MRBW without the path-consistency (PC) algorithm; it proposes the best rule for each recipe by counting the occurrences and picking the most frequent one. 
2) TILR-PC: this mode is the same as MRBW but with the PC algorithm implemented.
3) $\text{TILR}_{\bm{\theta}}$-PC: this mode performs MRBW and PC algorithm during search and learns rule via the parametric model introduced in \S\ref{sec:learning-model}.

We compare TILR with baseline methods that reason on the clique-expansion graphs (as TILR is the only method that reasons on hypergraphs): 
1) GNN-GCN~\citep{kipf2016semi} and GNN-Cheb~\cite{defferrard2016convolutional}.
GNN methods are evaluated on the clique-expansion graph with the time labels.
Specifically, we set GNN methods to the inductive setting, where the entities all share the same one-hot vector. 
This way GNNs learn to generalize to classify graphs with unseen entities. 
2) CTDNE~\citep{nguyen2018continuous}.
A graph embedding method that utilizes the time point labels in the clique-expansion graph. 
3) traditional ILP methods: 
random walk (RW),
NeuralLP~\citep{yang2017differentiable}, 
NLIL~\citep{yang2020learn}, 
and Drum~\citep{sadeghian2019drum}. 
We implement the RW method as a simplistic baseline that performs personalized random walks on the clique-expansion graph while ignoring the time labels; it counts the frequency of paths that answer the query. 
All experiments are done on a PC with i7-8700K and one GTX1080ti. We use the Mean Reciprocal Rank (MRR) and Hits@3 and Hits@10 as the metrics.

\begin{table*}[t]
\centering
\resizebox{0.95\textwidth}{!}{%
\begin{tabular}{@{}lcccccclcccccc@{}}
\toprule
\multirow{3}{*}{Method}        & \multicolumn{6}{c}{YouCook2-HG}                                                               &  & \multicolumn{6}{c}{nuScenes-HG}                                                               \\ \cmidrule(lr){2-7} \cmidrule(l){9-14} 
                               & \multicolumn{3}{c}{Clique-Expansion}          & \multicolumn{3}{c}{Hypergraph}                &  & \multicolumn{3}{c}{Clique-Expansion}          & \multicolumn{3}{c}{Hypergraph}                \\
                               & MRR           & Hits@3        & Hits@10       & MRR           & Hits@3        & Hits@10       &  & MRR           & Hits@3        & Hits@10       & MRR           & Hits@3        & Hits@10       \\ \cmidrule(r){1-7} \cmidrule(l){9-14} 
GNN-GCN                        & 0.32          & 28.1          & 36.9          & -             & -             & -             &  & 0.40          & 32.2          & 50.4          & -             & -             & -             \\
GNN-Cheb                       & 0.37          & 32.7          & 43.7          & -             & -             & -             &  & 0.41          & 32.6          & 55.6          & -             & -             & -             \\
CTDNE                          & 0.15          & 11.9          & 16.6          & -             & -             & -             &  & 0.08          & 9.7           & 11.3          & -             & -             & -             \\ \cmidrule(r){1-7} \cmidrule(l){9-14} 
RW                             & 0.19          & 18.8          & 20.6          & -             & -             & -             &  & 0.15          & 11.0          & 17.4          & -             & -             & -             \\
NeuralLP                       & 0.36          & 35.4          & 39.5          & -             & -             & -             &  & 0.47          & 39.4          & 63.1          & -             & -             & -             \\
NLIL                           & 0.38          & 36.3          & {\ul 46.8}    & -             & -             & -             &  & 0.45          & 38.3          & 58.2          & -             & -             & -             \\
Drum                           & \underline{0.40}    & \underline{38.5}    & 45.2          & -             & -             & -             &  & \underline{0.51}    & \underline{45.5}    & \textbf{66.8} & -             & -             & -             \\ \cmidrule(r){1-7} \cmidrule(l){9-14} 
TILR                           & 0.25          & 20.4          & 23.7          & 0.35          & 29.4          & 36.3          &  & 0.14          & 10.1          & 19.5          & 0.19          & 20.8          & 31.4          \\
TILR-PC                        & 0.30          & 27.9          & 31.1          & \underline{0.60}    & \underline{55.8}    & \underline{59.1}    &  & 0.18          & 15.1          & 22.9          & \underline{0.22}    & \underline{24.1}    & \underline{35.5}    \\
$\text{TILR}_{\bm{\theta}}$-PC & \textbf{0.44} & \textbf{40.2} & \textbf{47.3} & \textbf{0.72} & \textbf{76.0} & \textbf{79.4} &  & \textbf{0.52} & \textbf{47.2} & \underline{65.2}    & \textbf{0.64} & \textbf{66.0} & \textbf{81.1} \\ \bottomrule
\end{tabular}%
}
\caption{MRR and Hits of TILR and baselines on YouCook2 recipe summarization and nuScenes behavior explanation benchmarks.}
\label{tab:main-res}
\vspace{-.3cm}
\end{table*}

\begin{table}[h]
\centering
\resizebox{0.9\linewidth}{!}{%
\begin{tabular}{@{}ll@{}}
\toprule
Recipe               & Example temporal rules                                                                                                                                                                    \\ \midrule
BLT Sandwich         & \begin{tabular}[c]{@{}l@{}}Put(Mayo, Bread1) \\ \; BEFORE MixInto((Bread1, lettuce, tomato), M) \\ \; BEFORE Cover(Bread2, M)\end{tabular}                                                \\ \midrule
Onion Ring           & \begin{tabular}[c]{@{}l@{}}Remove(Layers, Onion)\\ \; AND Cut(Onion, Onion)\\ \; BEFORE MixInto((Onion, flour), M)\\ \;  BEFORE Put(M, Fryer)\end{tabular}                                \\ \midrule
Spaghetti \& Meatball & \begin{tabular}[c]{@{}l@{}}MixInto((Garlic, Beef), M1)\\ \; BEFORE Saute(M1, M1)\\ \;  AND MixInto((Tomato Paste, cheese), M2)\\ \;  BEFORE MixInto((M1, M2, Spaghetti), M3)\end{tabular} \\ \bottomrule
\end{tabular}%
}
\caption{Example rules that summarize the recipes. For simplicity, we simplify the FOL notations $Put(X, Y) \land Mayo(X) \land Bread(Y)$ into $Put(Mayo, Bread1)$, where we use digits $Bread1$, $Bread2$ to distinguish different entities of the same class. The class $M$ denotes the intermediate products in the cooking process.}
\label{tab:recipe-examples}
\vspace{-.3cm}
\end{table}

\textbf{Results}.
The results are shown in Table~\ref{tab:main-res}.
We find TILR yields the best performance on the hypergraphs, which is 30\% higher than the ordinary graph counterparts.
This suggests the proposed temporal hypergraph representation can better capture the higher-order information, which addresses question \textbf{(Q1)}.
On the other hand, we also find TILR outperforms all baseline methods on clique-expansion graphs as well.
This suggests that TILR, with the introduction of time intervals, generalizes better than other baselines in the temporal domain, as others can at most utilize time point information.
To further inspect the generalizability of TILR, we show example rules in Table~\ref{tab:recipe-examples}.
This shows that TILR can generalize meaningful and interpretable rules over both logic and temporal domains, addressing question \textbf{(Q2)}.
We investigate \textbf{(Q3)} in \S\ref{sec:exp-nuscenes} with both tasks' results.



\subsection{Driving behavior explanation}
\label{sec:exp-nuscenes}

The nuScenes autonomous driving dataset~\citep{caesar2020nuscenes} consists of 750 driving scenes in the training set. Each scene is a 20s video clip; each frame in the clip is annotated with 2D bounding boxes and class labels. The dataset also provides lidar and ego information such as absolute position, brake, throttle, and acceleration.

\begin{table}[h]
    \centering
    \resizebox{0.9\linewidth}{!}{%
        \begin{tabular}[b]{@{}llcccc@{}}
        \toprule
        \multicolumn{2}{l}{\multirow{2}{*}{Predicate Type}} & \multirow{2}{*}{\# Predicates} & \multicolumn{2}{c}{\# Facts}                     & \multirow{2}{*}{Examples} \\
        \multicolumn{2}{l}{}                                &                                & Per Graph             & Total                    &                           \\ \midrule
        \multirow{4}{*}{Relation} & \multirow{2}{*}{Unary}  & \multirow{2}{*}{13}            & \multirow{2}{*}{4631} & \multirow{2}{*}{3473257} & ego.throttle              \\
                                  &                         &                                &                       &                          & ego.brake                 \\
                                  & \multirow{2}{*}{Binary} & \multirow{2}{*}{6}             & \multirow{2}{*}{1859} & \multirow{2}{*}{1394253} & in\_front                 \\
                                  &                         &                                &                       &                          & behind                    \\
                                  & N-ary                   & 1                              & 1692                  & 1269004                  & Between                   \\ \midrule
        \multicolumn{2}{c}{\multirow{3}{*}{Class}}          & \multirow{3}{*}{23}            & \multirow{3}{*}{257}  & \multirow{3}{*}{45881}   & vehicle.car               \\
        \multicolumn{2}{c}{}                                &                                &                       &                          & human.ped.adult    \\
        \multicolumn{2}{c}{}                                &                                &                       &                          & ped\_crossing             \\ \bottomrule
        \end{tabular}%
        }
    \caption{Stats. of the hypergraph benchmark \textbf{nuScenes-HG}.}
    \label{tab:auto-drive}
    \vspace{-.3cm}
\end{table}

\textbf{nuScenes-HG construction}.
We construct the hypergraph for each driving scene.
We use the 23 object classes as class labels, including \texttt{vehicle.car} and \texttt{human.pedestrian.adult};
We convert ego information and attributes such as \texttt{vehicle.moving} and \texttt{pedestrian.moving} into unary predicates.
The original dataset does not provide relations for the objects.
Here, we create the relative spatial relations by inferring from the absolute spatial information.
This includes \texttt{In\_front}, \texttt{Behind}, \texttt{Between} and etc.
The statistics are shown in Table~\ref{tab:auto-drive}.
Clique-expansion graphs are created for \textbf{nuScenes-HG} the same way as that in \textbf{YouCook2-HG}.

\textbf{Task}.
We apply TILR to explain the driving behavior with temporal logic rules.
Such an explanation is beneficial as it can provide insight into why certain behavior happens, for example, one can explain in what situation will the driver hit the brake; the evidence is likely to be there is a pedestrian crossing the street.
Formally, we consider the positive queries as events of \texttt{ego.brake} and \texttt{ego.throttle} and the rest of the events as the negative queries. 
Let $\rve = P\brB{\vx^0, \vx^n, \vt}$ be the event and $\cG$ be the temporal graph, we have
\begin{align*}
    & \cQ_+ = \{\rve | P \in \{ \texttt{ego.brake}, \texttt{ego.throttle}\}, \rve\in\cG \}, \\
    & \cQ_- = \{\rve | P \notin \{ \texttt{ego.brake}, \texttt{ego.throttle}\}, \rve\in\cG\}.
\end{align*}

\textbf{Results}.
The results are shown in Table~\ref{tab:main-res}.
In general, TILR performs the best on both clique-expansion graphs and hypergraphs, which is similar to that in \textbf{YouCook2-HG}, suggesting temporal hypergraph and the proposed TILR are the better representation and ILP method for learning rules from real-world data with temporal and higher-order relations.

\textbf{Ablation study}.
For both benchmarks, the PC algorithm and parameter model improve the performance consistently, suggesting they are essential components of TILR, resolving question \textbf{(Q3)}.
Furthermore, we find that 
$\text{TILR}_{\bm{\theta}}$-PC performs significantly better than two other modes in this benchmark
This is because the driving scenes are more diverse and many situations can lead to the target event, which makes it difficult to learn rules by counting. 
In this case, a learned model that proposes the rule for individual scenes yields a higher score.

\vspace{-.2cm}
\section{Conclusion}

We propose temporal inductive logic reasoning (TILR), a novel framework that extends the ILP technique to applications beyond KGs. We introduce the temporal hypergraph and show that it is a powerful representation for many applications; we then propose the multi-start random B-walk (MRBW) algorithm, which, combined with the path-consistency algorithm, serves as the foundation of TILR.
In experiments, we create and release two dedicated temporal hypergraph datasets and evaluate TILR with strong baselines, which shows superior reasoning capabilities.

\section{Acknowledgement}

This work was supported by a sponsored research award by Cisco Research.

\bibliographystyle{named}
\bibliography{ijcai24}


\appendix

\section{Hypergraphs}
\label{app:hypergraph-discussion}

In this work, we develop learning algorithms over B-graphs. 
While B-graphs are a subset of all hypergraphs, it represents a wide range of hypergraphs and many other types can be either converted to B-graph or solved by straightforwardly extending our proposed algorithm.
All directed hypergraphs can be categorized as B-graphs, F-graphs, or BF-graphs~\citep{gallo1993directed}.
In particular, F-graphs are the opposite of B-graphs with one head and multiple tails, where the proposed MRBW can be applied without modification, and BF-graphs can be converted to B-graphs straightforwardly for random walk purposes~\citep{gallo1993directed}.

\section{Example Hypergraphs}
\label{app:example-hypergraphs}

We show example temporal hypergraphs of recipes
\ttt{Onion Rings},
\ttt{Shepherd's Pie},
and \ttt{Tuna Sashimi}
in Figure~\ref{fig:temporal-hypergraph-examples}.

\section{Experiments}\label{app:exp-details}

\textbf{Recipe summarization}.
The goal of recipe summarization is to generalize a structured procedure for each distinct recipe from the hypergraph data as that in Figure~\ref{fig:example-graph}. We formalize this problem as a \textit{graph classification} task.
However, treating it as a simple classification problem is not sufficient for generating rules for summarization. Without additional constraints, solving ~\eq{eq:recipe-background-pos-negs} can lead to rules with good accuracy but poor interpretability. For example, one can learn rules that classify \textit{spaghetti $\&$ meatballs} easily by solely checking if the \texttt{Spaghetti} and \texttt{Meatball} class labels are present in the graph. 
To solve this, we put the constraint that the rules must cover most of the time span of the target hypergraphs. 
In other words, it needs to find rules that generalize to a significant subgraph in the hypergraph. 
This can be done by keeping track of the earliest and the latest events when a subgraph is matched by a rule. If the duration of the subgraph is less than that of the entire graph then the rule is discarded.

\clearpage
\newpage
\begin{minipage}[t]{0.98\textwidth}
    \centering
    \begin{minipage}[t]{\textwidth}
        \centering
        \includegraphics[width=\linewidth, trim={8cm 0 6cm 0},clip]{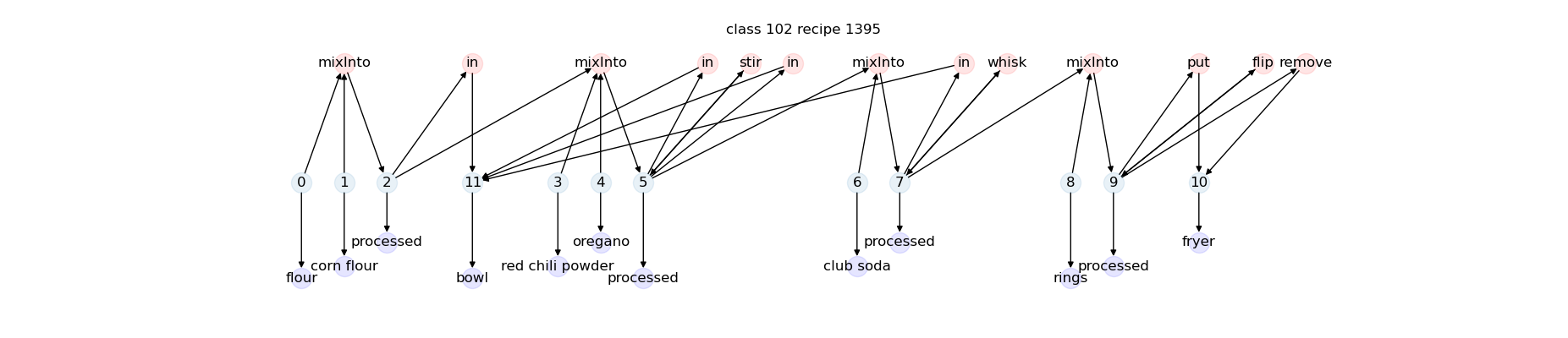}
    \end{minipage}
    \begin{minipage}[t]{\textwidth}
        \centering
        \includegraphics[width=\linewidth, trim={8cm 0 6cm 0},clip]{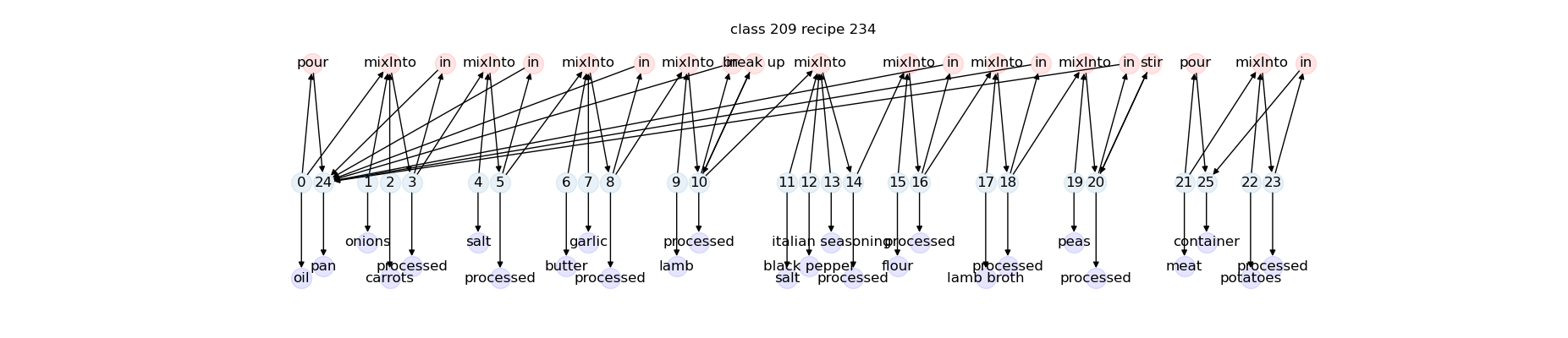}
    \end{minipage}
    \begin{minipage}[t]{\textwidth}
        \centering
        \includegraphics[width=\linewidth, trim={8cm 0 6cm 0},clip]{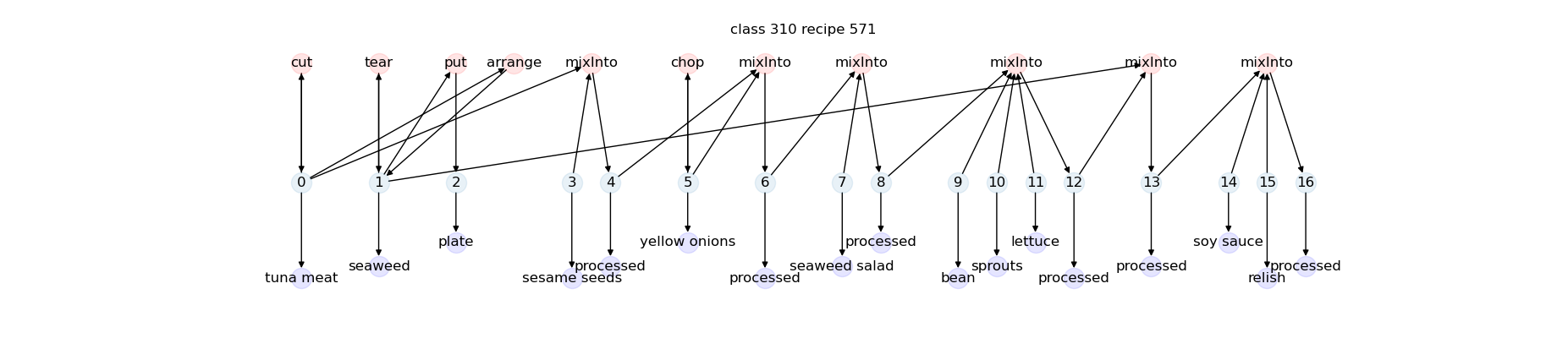}
    \end{minipage}
    \captionof{figure}{Example temporal hypergraphs for recipes Onion rings, Shepherd's Pie, and Tuna Sashimi.}
    \label{fig:temporal-hypergraph-examples}
\end{minipage}

\end{document}